\documentclass[runningheads]{llncs}
\usepackage[T1]{fontenc}
\usepackage{float}
\usepackage{graphicx}
\usepackage{hyperref}
\usepackage{float}
\hypersetup{
    colorlinks=true,
    linkcolor=blue,
    filecolor=magenta,   
    citecolor=blue,
    urlcolor=cyan,
    pdftitle={EHRmonize},
    pdfpagemode=FullScreen,
    }

\usepackage{subcaption}
\usepackage{color, colortbl}
\usepackage{multirow}
\newcolumntype{P}[1]{>{\centering\arraybackslash}p{#1}}
\definecolor{Gray}{gray}{0.93}

\begin{document}
\title{Evaluating the Impact of Pulse Oximetry Bias in Machine Learning under Counterfactual Thinking}
\titlerunning{Impact of Pulse Oximetry Bias in Machine Learning}
%
\author{Inês Martins\inst{1,2}\thanks{Corresponding Authors: \email{\{ines.a.martins,jaime.cardoso\}@inesctec.pt}} 
\and
João Matos\inst{3,4}
\and
Tiago Gonçalves\inst{1,2}
\and
Leo A. Celi\inst{4}
\and
An-Kwok Ian Wong\inst{3}
\and
Jaime S. Cardoso\inst{1,2}$^{\star}$}

\authorrunning{I. Martins et al.}

\institute{Faculty of Engineering, University of Porto\and Institute for Systems and Computer Engineering, Technology and Science \and Duke University\and Massachusetts Institute of Technology}

\maketitle              
\begin{abstract}
Algorithmic bias in healthcare mirrors existing data biases. However, the factors driving unfairness are not always known. Medical devices capture significant amounts of data but are prone to errors; for instance, pulse oximeters overestimate the arterial oxygen saturation of darker-skinned individuals, leading to worse outcomes. The impact of this bias in machine learning (ML) models remains unclear. This study addresses the technical challenges of quantifying the impact of medical device bias in downstream ML. Our experiments compare a ``perfect world'', without pulse oximetry bias, using SaO$_2$ (blood-gas), to the ``actual world'', with biased measurements, using SpO$_2$ (pulse oximetry). Under this counterfactual design, two models are trained with identical data, features, and settings, except for the method of measuring oxygen saturation: models using SaO$_2$ are a ``control'' and models using SpO$_2$ a ``treatment''. The blood-gas oximetry linked dataset was a suitable test-bed, containing 163,396 nearly-simultaneous SpO$_2$ - SaO$_2$ paired measurements, aligned with a wide array of clinical features and outcomes. We studied three classification tasks: in-hospital mortality, respiratory SOFA score in the next 24 hours, and SOFA score increase by two points. Models using SaO$_2$ instead of SpO$_2$ generally showed better performance. Patients with overestimation of O$_2$ by pulse oximetry of $\geq$ 3\% had significant decreases in mortality prediction recall, from 0.63 to 0.59, \textit{P} < 0.001. This mirrors clinical processes where biased pulse oximetry readings provide clinicians with false reassurance of patients' oxygen levels. A similar degradation happened in ML models, with pulse oximetry biases leading to more false negatives in predicting adverse outcomes.

\keywords{Bias \and Machine Learning \and Medical Devices \and Pulse Oximetry}
\end{abstract}
%
%
\section{Introduction}
Machine learning (ML) has the potential to revolutionize healthcare, promising increased objectivity in decisions, enhanced health system efficiency, and better overall health outcomes~\cite{balagopalan2024machine}. However, effective deployment of ML applications in healthcare is happening at a slower pace than expected. Obermeyer and colleagues' seminal work in 2019~\cite{obermeyer2019dissecting} raised concerns about the risk of bias in health ML, highlighting that an algorithm widely used in U.S. hospitals was less likely to refer Black people than White people (with similar illnesses) to more personalized treatment programs. These inconsistencies were attributed to the fact that the model was based on health care cost, instead of actual illness~\cite{obermeyer2019dissecting}.

However, understanding the causality and underlying factors driving downstream unfairness in ML models for healthcare is challenging due to significant confounding variables. Despite this complexity, addressing these issues is crucial for mitigating biases and improving model performance. Medical devices, such as pulse oximeters, thermometers, and sphygmomanometers, may introduce similar inconsistencies in model results due to calibration flaws~\cite{charpignon2023critical}. These devices are routinely used to collect vital signs to support clinical decision-making, especially in the Intensive Care Unit (ICU), where patients are more unstable \cite{Sauer2022}.

Pulse oximeters estimate arterial oxygen saturation by measuring light absorption at two light wavelengths (660nm - red and 940nm - infrared) of oxyhemoglobin and deoxyhemoglobin in capillary blood. However, this physical principle can be independently affected by skin tone~\cite{hao2024utility,Jubran2015}. Moreover, it is known that oximeters' original validation was not performed on a diverse population~\cite{moran2020popular}. Literature provides evidence that these devices measure the blood oxygen saturation differently across subpopulations~\cite{moran2020popular,sjoding2020racial}. Sjoding et al.~\cite{sjoding2020racial} found that Black patients experienced nearly triple hidden hypoxemia cases compared to White patients when using pulse oximetry measurements (SpO$_2$) instead of arterial oxygen saturation in arterial blood gas (SaO$_2$). These discrepancies were associated with inequities in oxygen therapies, subsequently higher organ dysfunction scores, and increased mortality rates among subpopulations~\cite{wong2021analysis}. And still, existing devices are likely to keep being used in a myriad of environments while no better devices are developed and regulated~\cite{dempsey2023high}.

Although the impact of pulse oximetry bias on patient outcomes is well-documented, its effect on downstream ML models using these biased measurements remains unknown. This study aims to address the question: \textit{How can we assess whether a model's performance and fairness are affected by a feature encoding racial bias?} Utilizing counterfactual thinking and the pulse oximetry use case, we aim to develop a framework to evaluate the impact of medical device bias on downstream ML tasks. The main contributions of this paper are:

\begin{itemize}
\item \textbf{Counterfactual approach:} We introduce a novel methodological framework that leverages counterfactual thinking to analyze the impact of medical device bias on downstream machine learning performance and fairness across subgroups;
\item \textbf{Comprehensive evaluation:} We conduct a wide array of experiments using the pulse oximetry bias use case, across three different clinical prediction tasks, and two different ML models. We utilize the blood-gas linked dataset (BOLD)~\cite{matos2024bold} as a test-bed, including data from MIMIC-III~\cite{johnson2016mimic}, MIMIC-IV~\cite{johnson2023mimic}, and eICU-CRD~\cite{pollard2018eicu}. 
\end{itemize}

Our work can be extended to other medical devices and databases, providing a useful approach for researchers examining algorithmic bias in health ML. The code developed in this study is publicly available in a GitHub repository\footnote{\url{https://github.com/InesAMar/PulseOxBias}}.

\section{Methodology}\label{sec:methods}
\subsection{Counterfactual approach}
\begin{figure}[ht]
    \centering
    \includegraphics[width=0.9\linewidth]{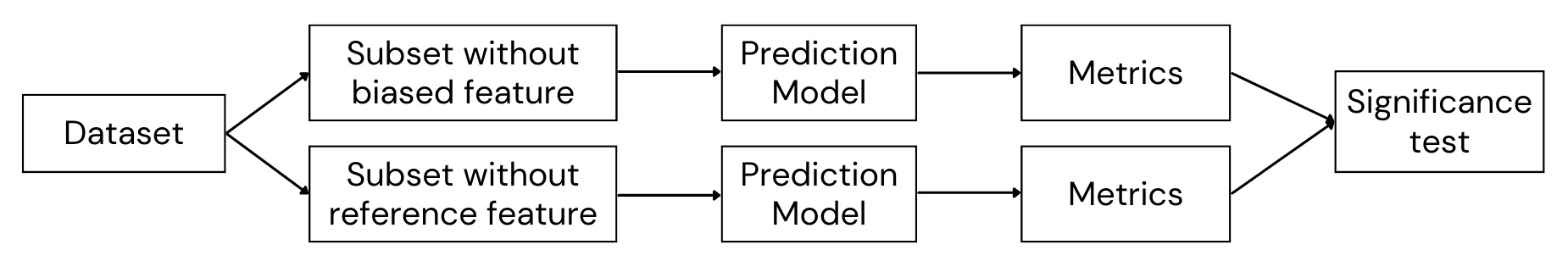}
    \caption{Assessment of the impact of medical device bias on downstream ML. 
    }
    \label{fig:model_structure_v2}
\end{figure}

We designed our experiments to compare a ``perfect world'', where pulse oximetry bias does not exist - a measure that SaO$_2$, the blood-gas reading, can provide - to the ``actual world'', where pulse oximetry bias affects certain subgroups of patients - as measured by SpO$_2$. Under this counterfactual design, the former can be interpreted as a ``control'', and the latter as the ``treatment''. In a ML setting, this necessitates keeping all other variables (``confounders'') - train and test split; remaining features; classification task; and evaluation metric - the same for both groups, ensuring that the device used to measure arterial O$_2$ saturation is the sole varying factor (Figure~\ref{fig:model_structure_v2}).

Finding ``counterfactuals'' in real-world data is particularly challenging in the medical domain because no two patients are inherently the same, making it difficult to account for confounding factors. Therefore, we employed BOLD - a blood-gas and oximetry linked dataset - where SaO$_2$ and SpO$_2$ are paired per patient. As each patient of this dataset has both measures nearly simultaneously, we can control for all other variables and assess the impact of pulse oximetry bias for each patient, in terms of ML performance.

\subsection{Dataset}
BOLD was created by harmonizing three Electronic Health Record databases (MIMIC-III, MIMIC-IV, eICU-CRD), comprising ICU stays of U.S. patients~\cite{matos2024bold}. It contains paired pulse oximetry readings (SpO$_2$) and preceding arterial blood gas measurements (SaO$_2$), acquired within a 5-minute interval. Pairs with values between 70\% and 100\% are included. Patient characteristics, vital signs, laboratory values, and  Sequential Organ Failure Assessment (SOFA) scores \cite{jones2009sequential} are time-aligned with the SaO$_2$ sample. To augment our sample size for this study, we extended BOLD beyond the first pair of measurements per hospitalization, to include all possible pairs.

\subsection{Feature selection and Preprocessing}
The respiratory SOFA (rSOFA) score was computed by the difference between the overall SOFA score and the sum of the remaining individual scores (coagulation, liver, cardiovascular, central nervous system and renal). Features with clinical relevance were manually selected by a physician author (AIW)~\cite{hempel2023prediction,sun2023prediction,zhang2023development}. These included demographics: age and sex; comorbidities; vital signs: blood pressure, heart rate, respiratory rate, and temperature; laboratory test values: albumin, anion
gap, bicarbonate, blood urea nitrogen, creatinine, glucose, hemoglobin, lactate, platelet count, potassium,
red blood cell count, red cell distribution width, and sodium; and SOFA scores: overall, respiratory, and cardiovascular. Missing vital signs and laboratory values were imputed as the mid-point of the normal range~\cite{labvalues,2023shining}.

\subsection{Machine Learning}
Predictions were performed at the time of a (SpO$_2$, SaO$_2$), based on past features as described above and targeting outcomes in the next 24 hours. A stratified 10-fold cross-validation strategy (to divide between train and test sets) was performed to evaluate the model’s consistency across folds and improve the robustness of the results. This approach ensures that no patient would simultaneously be in both splits.
Three binary classification tasks were studied: in-hospital mortality, future rSOFA score (1 if $\geq$ 1 point, 0 otherwise), and future increased SOFA score (1 if increasing by at least two points~\cite{singer2016third}, 0 otherwise). Logistic regression (LR) and XGBoost classifier were fit and assessed. Their performance was evaluated with the area under the receiver operating curve (AUROC), recall, F1-score, and accuracy.

\subsection{Disparity axes and statistical inference}
The aforementioned evaluation metrics were computed across different subgroups of patients, according to different disparity axes:
\begin{itemize}
    \item \textbf{Race and ethnicity:} grouped into the following categories: ``Asian'', ``Black'', ``Hispanic or Latino'', ``White'' and ``Other or Unknown''. This variable is used as a surrogate for skin tone, the hypothesized root cause of bias \cite{hao2024utility}.
    \item \textbf{Magnitude of Bias:} grouped into four non-overlapping bins: < -3\%, between -3\% and 0\%, between 0\% and 3\%, and $\geq$ 3\%. This is a marker for patients who actually have faulty pulse oximetry readings and, therefore, are ``at risk'' (or not).
    \item \textbf{Hidden Hypoxemia (HH):} defined as having SaO$_2$ < 88\% and SpO$_2$ $\geq$ 88 \%~\cite{wong2021analysis}. This represents patients who are most affected by pulse oximetry bias.
\end{itemize}
Two-sided paired t-tests were performed to compare the performance of the SpO$_2$ vs. SaO$_2$ models, fold by fold.

\section{Results}\label{sec:results}
\subsection{Dataset description}
The extended version of BOLD contained 163,396 pairs, representing 34,252 patients described in Table \ref{tab:tableone_bold}. Positivity rates for the classification tasks revealed class imbalances: in-hospital mortality at 24.0\%; future rSOFA score at 41.6\%; and SOFA score increase at 23.8\%.

\begin{table}[t]
\centering
    \caption{Patient Characteristics of the Study Cohort (IQR, Interquartile Range).}
    \label{tab:tableone_bold}
\begin{tabular}{m{3cm}|P{1.65cm}|P{1.7cm}|P{1.7cm}|P{1.7cm}|P{1.8cm}}
\hline \rowcolor{Gray}
& \textbf{Asian} & \textbf{Black} & \textbf{Hispanic or Latino} & \textbf{Other} & \textbf{White} \\
n & 605 & 3397 & 1448 & 2823 & 25979 \\ \rowcolor{Gray}
Sex Female, n (\%) & 253 (41.8) & 1613 (47.5) & 661 (45.6) & 1142 (40.5) & 11251 (43.3) \\
Age, median [IQR] & 66.0 [54.0,78.0] & 61.0 [51.0,71.0] & 67.5 [53.0,78.0] & 65.0 [52.0,75.0] & 67.0 [57.0,77.0] \\ \rowcolor{Gray}
In-Hospital Mortality, n (\%) & 115 (19.0) & 589 (17.3) & 280 (19.3) & 501 (17.7) & 4528 (17.4) \\
Hidden Hypoxemia, n (\%) & 15 (2.5) & 129 (3.8) & 40 (2.8) & 71 (2.5) & 722 (2.8) \\ \rowcolor{Gray}
Comorbidity Score, median [IQR] & 4.0 [2.0,6.0] & 4.0 [2.0,6.0] & 4.0 [2.0,6.0] & 4.0 [2.0,6.0] & 4.0 [2.0,6.0] \\
SOFA Past Overall 24hr, median [IQR] & 5.0 [2.0,8.0] & 5.0 [3.0,8.0] & 5.0 [3.0,8.0] & 6.0 [3.0,8.0] & 5.0 [3.0,8.0] \\ \rowcolor{Gray}
SOFA Future Overall 24hr, median [IQR] & 4.0 [2.0,7.0] & 5.0 [3.0,8.0] & 5.0 [3.0,7.0] & 5.0 [3.0,8.0] & 5.0 [3.0,7.0] \\
N pairs (per hosp. adm.), median [IQR] & 2.0 [1.0,5.0] & 2.0 [1.0,5.0] & 2.0 [1.0,6.0] & 2.0 [1.0,5.0] & 2.0 [1.0,5.0] \\
\end{tabular}
\end{table}

\subsection{Experiments}

\textbf{Different model architectures:} The two model architectures had similar results, but XGBoost generally outperformed LR. Only XGBoost results will be presented in detail.

\begin{figure}[ht]
    \centering
    \includegraphics[width=1\linewidth]{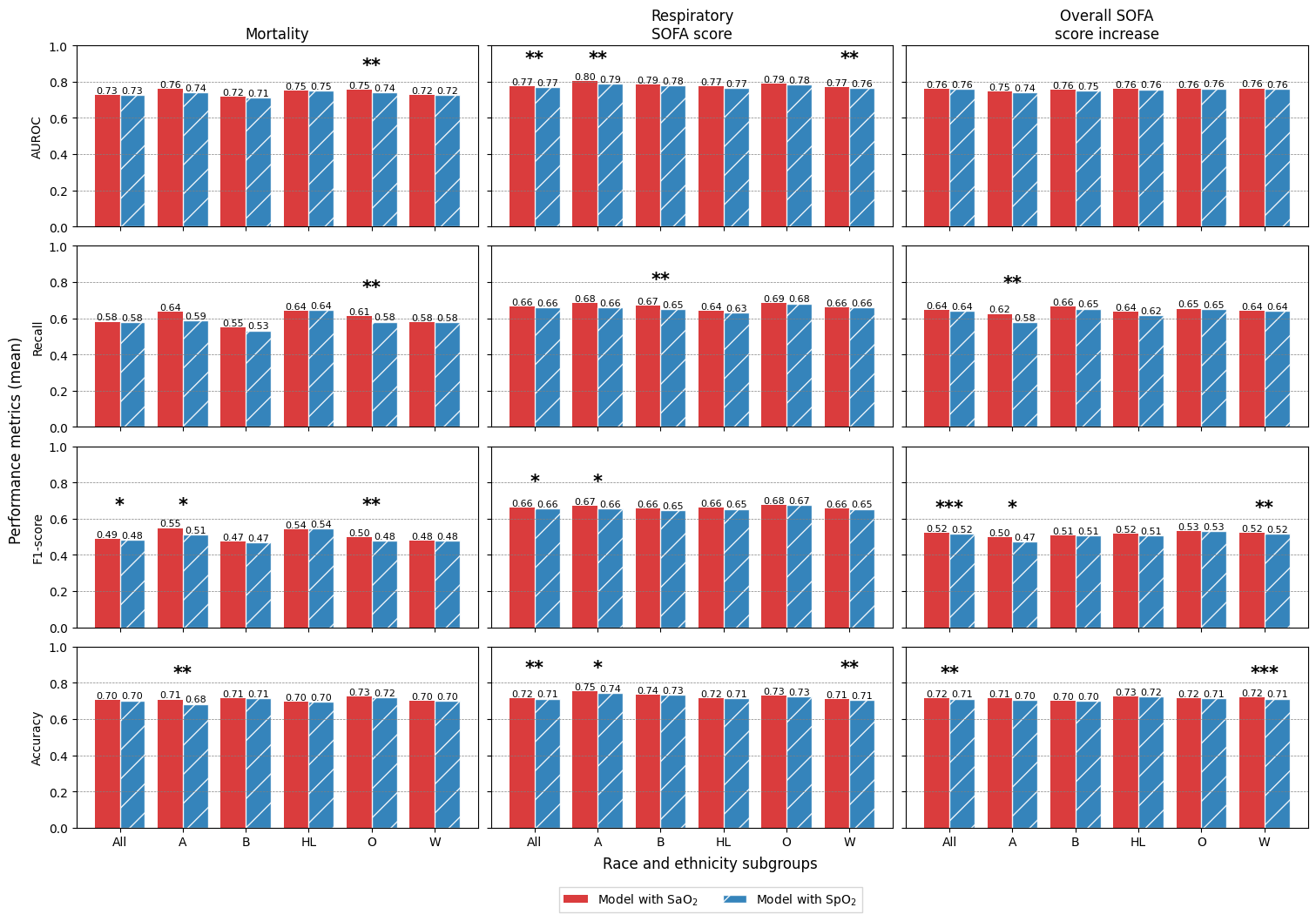}
    \caption{Mean value of the XGBoost performance metrics across race and ethnicity subgroups. Significant differences between SpO$_2$ and SaO$_2$ models are identified with: ``$\ast$'', for p-values $\leq$ 0.05; ``$\ast\ast$'', for p-values $\leq$ 0.01; or ``$\ast\ast\ast$'', for p-values $\leq$ 0.001. A: Asian; B: Black; HL: Hispanic or Latino; O: Other; W: White.}
    \label{fig:XGBClassifier0SpO2_BarChart_new1}
\end{figure}

\textbf{Across racial groups:} By clustering the results across race and ethnicity groups, several situations were identified as significantly different by the XGBoost model, as shown in Figure \ref{fig:XGBClassifier0SpO2_BarChart_new1}. For example, Asian patients had a degradation from 0.55 to 0.51 in F1-Score (\textit{P} < 0.05), when using SpO$_2$ as a feature, a trend that was verified across metrics and tasks for these patients.

\begin{figure}[ht]
\centering
    \includegraphics[width=1\linewidth]{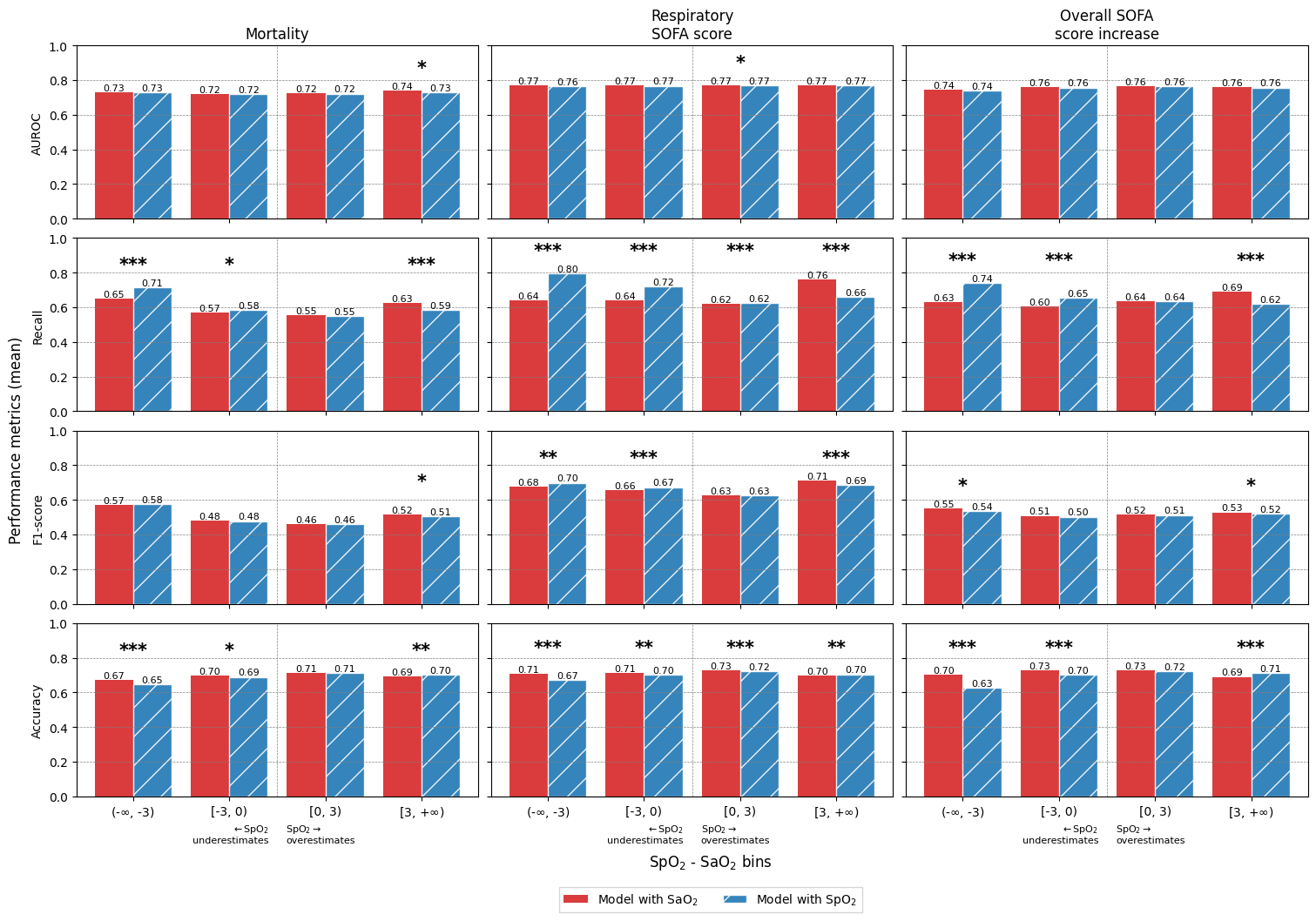}
    \caption{Mean value of the XGBoost performance metrics across disparity groups. Significant differences between SaO$_2$ and SpO$_2$ models are identified with: ``$\ast$'', for p-values $\leq$ 0.05; ``$\ast\ast$'', for p-values $\leq$ 0.01; or ``$\ast\ast\ast$'', for p-values $\leq$ 0.001.}
    \label{fig:XGBClassifier1SpO2_BarChart_new1}
\end{figure}

\textbf{Across magnitude of bias:} Figure \ref{fig:XGBClassifier1SpO2_BarChart_new1} represents the bias effect on XGBoost prediction performance when disparities were divided into four bins, according to the difference between SpO$_2$ and SaO$_2$ values.
Accuracy is significantly higher in the SaO$_2$ model when SpO$_2$ underestimates O$_2$, and vice-versa. On the other hand, recall is significantly higher in the SaO$_2$ model when SpO$_2$ overestimates O$_2$, and vice-versa. Differences in performance are exacerbated by higher disparities. 

\begin{figure}[ht]
    \centering
    \includegraphics[width=0.7\linewidth]{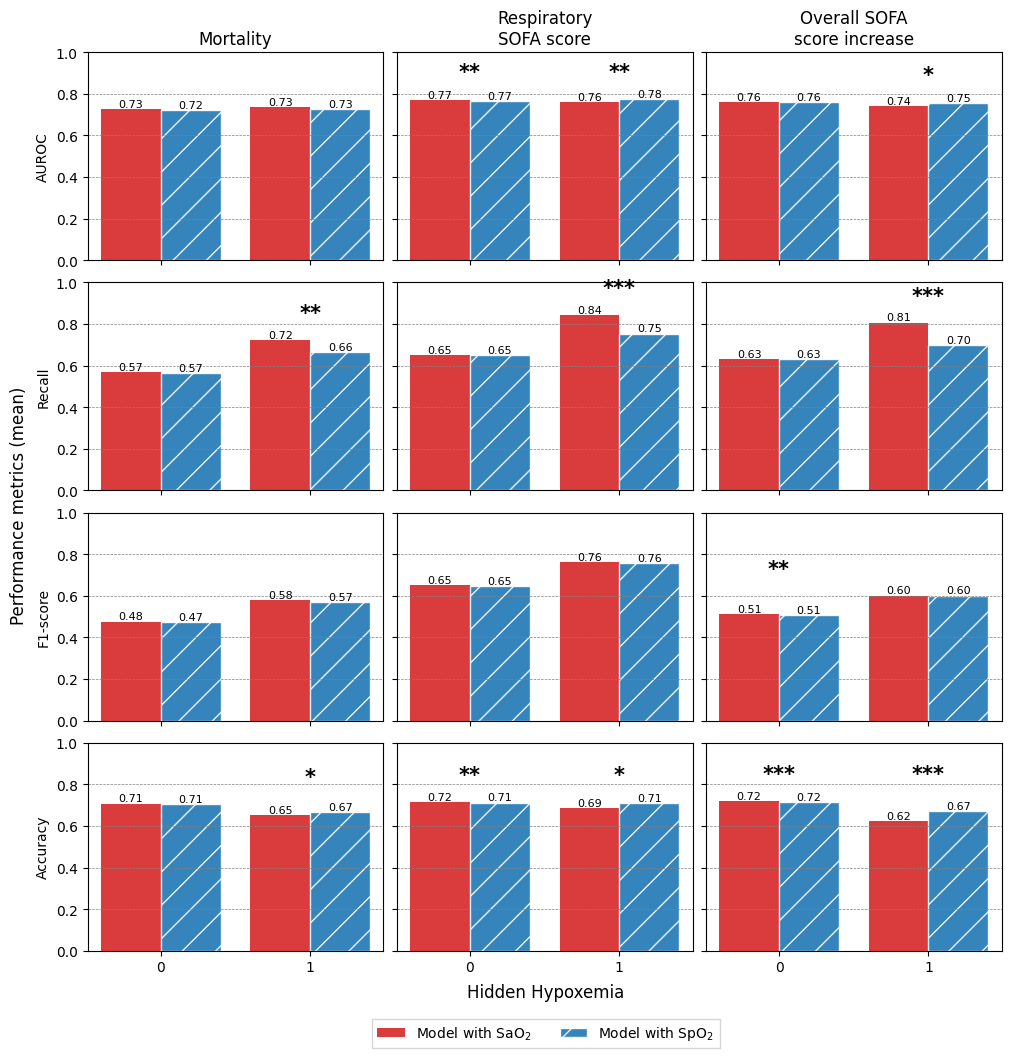}
    \caption{Mean value of the XGBoost performance metrics between patients with consistent SaO$_2$ and SpO$_2$ values (above or equal to 88\%) - class 0 - and the ones with hidden hypoxemia - class 1. Significant differences between SaO$_2$ and SpO$_2$ models are identified with: ``$\ast$'', for p-values $\leq$ 0.05; ``$\ast\ast$'', for p-values $\leq$ 0.01; or ``$\ast\ast\ast$'', for p-values $\leq$ 0.001.}
    \label{fig:XGBClassifier2SpO2_BarChart_new1}
\end{figure}

\textbf{Across HH groups:} Results show that accuracy is significantly lower and recall is significantly higher to the SaO$_2$ model in patients with HH (Figure \ref{fig:XGBClassifier2SpO2_BarChart_new1}). These findings agree with the ones above. In the class 0 group, significant differences were associated with higher performance of the SaO$_2$ model, which is in line with the results from the full cohort (Figure \ref{fig:XGBClassifier0SpO2_BarChart_new1}).

The aforementioned trends are consistent across tasks, where groups that are more ``at risk'' exhibit ML performance degradation in models using SpO$_2$, as opposed to SaO$_2$.
AUROC results often present marginal differences in performance. However, the model with SaO$_2$ usually attains higher values.
Results in Figures \ref{fig:XGBClassifier0SpO2_BarChart_new1}, \ref{fig:XGBClassifier1SpO2_BarChart_new1} and \ref{fig:XGBClassifier2SpO2_BarChart_new1} can be significantly different where the means appear equal because the performance metrics means were rounded to 2 decimal places.

\section{Discussion and Conclusion}\label{sec:discussion}
This work presents a counterfactual approach to quantify the impact of medical device bias in ML performance. Empirically applied to the pulse oximetry use case, we compare two otherwise identical ML models, differing only in their use of either unbiased SaO$_2$ (blood-gas) data or biased SpO$_2$ (pulse oximetry) data. By isolating the method of O$_2$ measurement, we directly target the quantification of bias. Evaluated using BOLD (N = 163,396), we conducted experiments on three binary classification tasks.

In our ML tasks, the positive classes correspond to adverse health outcomes, while the negative class represents a less sick population. O$_2$ overestimation by SpO$_2$ ([3, +$\infty$) disparity group in Figure \ref{fig:XGBClassifier1SpO2_BarChart_new1}) may increase correct predictions for the negative class, contributing to higher accuracy. However, this also leads to a decrease in the correct predictions for the positive class and, consequently, in recall. Misidentifying positive cases due to biased measurements from medical devices could exacerbate disparities in healthcare diagnoses and treatment access. HH was higher for Black patients - 3.8\% (see Table \ref{tab:tableone_bold}), suggesting that models with SpO$_2$ would have worse performance. Although differences were verified, they were not statistically significant. An additional analysis would have to be performed across race and ethnicity groups.
The experiments across racial and ethnic groups did not show systematic differences in the remaining groups, as expected. In fact, using race and ethnicity as a disparity axis is debatable, as it is often considered a ``social construct'', and herein used an imperfect surrogate for skin tone, the hypothesized root cause of bias \cite{hao2024utility}. The analyses across the degree of bias and HH present more occurrences of ML performance degradation.

Overall, our results mirror clinical scenarios where biased pulse oximetry readings provide clinicians with false reassurance of patients' oxygen levels. A similar degradation happened in ML models, with bias leading to more false negatives in predicting adverse outcomes. These results reinforce previous reports in the literature~\cite{fawzy2022racial,wong2021analysis}, suggesting that biased pulse oximetry readings not only lead to adverse outcomes and inequities in healthcare diagnoses but have the potential to exacerbate existing disparities if blindly fed to ML models.

The developed framework has the potential to be a valuable tool for advancing more just ML solutions in healthcare, revolving around bias and fairness. Its inherent counterfactual approach enhances the transparency and explainability of performance degradation across patient subgroups. Besides, our framework is easy to understand and use, as shown by our experiments with pulse oximetry data across three different clinical prediction tasks and two different machine learning algorithms. It brings a faster and more interesting approach for researchers dealing with algorithmic bias in health ML.
Additionally, given that this methodology is agnostic to the task and device, it can be easily applied to other use cases, such as temporal thermometry, which has also been found to exhibit racial bias due to similar underlying physical mechanisms (infrared light) being independently affected by skin pigmentation~\cite{bhavani2022racial}.

Our framework presents several limitations, as it depends on the availability of both a faulty and a gold-standard measurement. While BOLD provided a suitable test-bed, this may not be the case for all measurements from medical devices. Moreover, BOLD retrospectively aligns blood-gas and oximetry data, leveraging the vast amounts of information in Electronic Health Records, but this process can introduce errors and noise. Future work should explore the applicability of the presented methods in other settings to further validate their utility and versatility, potentially outside of healthcare, and investigate the integration of these methodologies in post-deployment AI. Naturally, we also expect this framework to contribute towards developing solutions to mitigate the effect of medical device bias in ML models within a real-world clinical setting.

\textbf{Prospect of application}:
Hospital systems should consider similar approaches before algorithm deployment. Further use cases need to be identified, as not all sources of bias allow for a counterfactual approach. It would be interesting to continue the study on medical devices that rely on light sensors, more specifically infrared light, such as temporal thermometers.

\begin{credits}
\subsubsection{\ackname} This work has received funding from the Portuguese Foundation for Science and Technology (FCT) through the Ph.D. Grant ``2020.06434.BD''.
\subsubsection{\discintname}
The authors have no competing interests in the paper.
\end{credits}

%
%
\bibliographystyle{splncs04}
\bibliography{mybibliography}

\end{document}